\documentclass{article}
\DeclareUnicodeCharacter{FF0C}{,}
\usepackage[margin=1in]{geometry}
\usepackage{titlesec}
\titleformat*{\section}{\large\bfseries}
\titleformat*{\subsection}{\bfseries}
\titleformat*{\subsubsection}{\itshape}

\usepackage{authblk}

\usepackage{amssymb, amsthm, mathtools} 
\usepackage{mleftright}
\usepackage{bm, bbm}

\usepackage{blindtext}

\usepackage{graphicx}
\usepackage{booktabs, float, multirow, threeparttable}
\usepackage[skip=5pt]{caption}
\usepackage[skip=5pt]{subcaption}
\usepackage{array}
\usepackage[detect-all]{siunitx}

\newrobustcmd{\B}{\DeclareFontSeriesDefault[rm]{bf}{b}\bfseries}

\usepackage{enumitem}
\setlist{nosep}

\usepackage{soul}
\usepackage{xcolor}

\usepackage{comment}

\usepackage[title]{appendix}

\usepackage[colorlinks, allcolors=blue]{hyperref}
\usepackage[nameinlink, capitalise]{cleveref}

\usepackage[
  backend=bibtex,
  style=authoryear,
  maxbibnames=3,
  minbibnames=3,
  maxcitenames=2,
  natbib=true
]{biblatex}
\title{Designing streetscapes from street-view imagery using diffusion models}

 \author[1]{Yuzhou Chen}
 \author[2]{Yuebing Liang}
 \author[3]{Lingqian Hu}
 \author[2]{Kailai Sun}
 \author[1]{Qingqi Song}
 \author[4]{Chang Zhao}
 \author[1]{Shenhao Wang*}

 \affil[1]{Department of Urban and Regional Planning, University of Florida}
 \affil[2]{Singapore-MIT Alliance for Research and Technology Centre (SMART)}
 \affil[3]{Department of Landscape Architecture and Urban Planning, Texas A\&M University}
 
 \affil[4]{Department of Agronomy, University of Florida}

 \date{}

\begin{document}
\allowdisplaybreaks
\maketitle
\vspace*{-2em}

\begin{abstract}
\noindent Street-view imagery (SVI) is widely used to quantify key indicators of urban environment, such as greenery, sky, or road view indices. However, existing studies largely focus on measuring current streetscapes and rarely support the generation of alternative and non-existing urban scenarios, which is a core task in geospatial disciplines such as urban planning and design. To address this gap, we propose a generative multimodal AI framework that synthesizes alternative streetscapes conditioned on targeted visual metrics, enabling direct visual exploration of urban scenarios. We first construct a multimodal dataset that aligns SVIs with textual descriptions, segmentation maps, road masks, and quantitative metrics of visual elements in Chicago and Orlando. Using this dataset, we demonstrate that diffusion models can produce realistic and semantically consistent streetscape imagery while responding to both textual and imagery controls. Our quantitative evaluations show that incorporating visual controls can improve semantic consistency, reducing the LPIPS index by approximately 6\% while maintaining global visual realism. In addition, overall semantic consistency increases by 23.7\% in Orlando and 46.4\% in Chicago, as measured by the mIoU index, with class-wise gains exceeding even 100\% improvement for building view indices. Streetscape generation can be controlled in a fine-grained manner by both visual and textual prompts, and when textual and visual controls conflict, imagery controls consistently dominate, indicating a clear control hierarchy and the importance of further developing visual controls for urban scene generation. Overall, this work establishes an important benchmark for streetscape generation using SVIs and diffusion models, and illustrates how generative AI can serve as a practical, scalable, and controllable approach for urban scenario exploration. 
\end{abstract}


\thispagestyle{empty}
\clearpage

\setcounter{page}{1}

\section{Introduction}
\label{sec:introduction}

\noindent Urban planners and designers frequently need tools that can generate alternative streetscape scenarios under specified design objectives \citep{batty2013new, gehl2013cities}. In many planning tasks, planners need to modify visual elements of the built environment while preserve others that are difficult to change. For example, a planner may wish to generate a new street corridor, which increases tree canopy coverage while preserves existing road geometry and building footprints. A designer may need to reduce the building view index along a dense urban corridor, while simultaneously increasing sky openness and the perceived spatial quality of the streetscape. In participatory planning, residents often need to compare alternative streetscapes under various greening policies, where computationally efficient streetscape generation is essential for meaningful engagement with quick feedback loops. Despite these needs, existing tools provide limited support for envisioning streetscapes under control and in a real-time manner, presenting an enduring challenge of translating abstract design intentions into concrete visual outcomes.

Street-view imagery (SVI) has emerged as a powerful data source for visualizing urban environments at the human eye level \citep{inoue_landscape_2022, ogawa_evaluating_2024}. It enables researchers to quantify visual elements such as greenery, sky exposure, buildings, and road infrastructure \citep{zhao_quantifying_2025}. These metrics have been widely used to study urban form, environmental quality, social equity, safety, active travel, and aesthetic values, directly informing design and planning practice \citep{biljecki2021street,li2022street,yang2009can, zhang2020building, weld2019deep}. However, existing SVI-based approaches remain fundamentally observational: they measure and evaluate current streetscape conditions but do not generate alternative or non-existing scenarios. This limitation is increasingly crucial as planning practice demands the ability to explore and compare visual alternatives aligned with specific design or planning principles, yet no computational tools exist to achieve this goal. Recent advances in diffusion models provide new opportunities for synthesizing realistic and controllable streetscapes. Compared to earlier generative approaches, diffusion models can produce high-quality and semantically consistent images conditioned on diverse inputs, including textual descriptions, spatial layouts, and segmentation maps \citep{ho2020denoising, ruiz2023dreambooth, inoue2023layoutdm}. These properties enable diffusion models to generate complex visual outcomes, particularly fitting streetscapes that need high-level realism and semantic consistency. Despite such advances, diffusion models have rarely been adopted in SVI-based urban and geospatial analysis. Existing generative studies in urban contexts have primarily focused on producing spatial maps, such as land-use maps or master plans, often using generative adversarial networks (GANs) \citep{ye_masterplangan_2022, wu_ganmapper_2022}. However, GANs generally suffer from training instability and limited controllability, and have been shown to produce less realistic outputs than diffusion models \citep{dhariwal2021diffusion}. In short, diffusion models have not yet been systematically integrated with SVIs to support the generation of realistic and controllable streetscapes. Moreover, prior work has largely overlooked the opportunities of leveraging visual and textual conditions to control the generative process, which is practically crucial because practitioners need mechanisms to translate their abstract intentions into visual street scenes. 

This study proposes to design streetscapes by using diffusion models to generate visually realistic SVIs. To achieve this goal, we firstly use semantic segmentation and object detection algorithms to extract quantitative visual elements of tree, building, sky, and road view indices, which are further embedded in textual descriptions. Further augmented with segmentation maps, our data processing leads to a multimodal data set that aligns SVI, quantitative visual metrics, textual descriptions, and segmentation masks. To demonstrate the generalizability of our work, we focus on two urban environments: Chicago, characterized by high-density development, and Orlando, known for its suburban sprawl. Empirically, we examine whether diffusion models can generate realistic and semantically consistent streetscapes by imposing textual and visual controls, as evaluated by indices such as Fréchet Inception Distance (FID), Learned Perceptual Image Patch Similarity (LPIPS), Structural Similarity Index (SSIM), and mean Intersection over Union (mIoU). We further vary the visual proportions of roads, sky, greenery, and buildings prespecified in the textual descriptions, and evaluate to what extent diffusion models can maintain visual realism and semantic consistency in out-of-sample scenarios. Lastly, we compare textual and imagery controls for streetscape generation, examining whether diffusion models can generalize across spatial contexts. Overall, we found that our streetscape generation is highly successful because the generated streetscapes are realistic, controllable, and semantically consistent with visual and textual prompts. 

As a result, this study offers three major contributions for generative geospatial analytics. First, we develop a multimodal dataset that aligns SVIs with textual descriptions, quantitative visual indicators, and semantic segmentation maps, enabling systematic implementation and evaluation of generative streetscape experiments.
Second, we present a diffusion framework for controllable streetscape generation, enabling the synthesis of realistic and semantically consistent SVIs using explicit visual and textual controls.
Lastly, we provide empirical evidence that streetscape generation is generalizable across urban contexts, offering insights into the interaction between textual and visual controls in generative geospatial tasks.

The paper is organized as follows. Section \ref{sec:literature_review} summarizes the past studies about SVIs and generative AI, and identifies the opportunities of generative AI for streetscape analysis. Section \ref{sec:methods_data} introduces multimodal data curation, diffusion models, and generative experiment design. Section \ref{sec:results} demonstrates that diffusion models can generate realistic, semantically consistent, and controllable streetscapes using prespecified visual and textual controls. Section \ref{sec:conclusion} summarizes our findings, future potentials, and limitations.

\section{Literature review}
\label{sec:literature_review}
\noindent Urban planning and design fields have long relied on visual representations to communicate design intentions and engage stakeholders. Early approaches to streetscape visualization depended on hand-drawn diagrams, physical models, and later computer-aided design tools, with researchers as early as the 1980s examining how visual simulations could be used to evaluate landscape and streetscape changes \citep{bechtel1987methods}. The advent of three-dimensional rendering software and virtual reality environments further advanced practitioners' ability to simulate proposed interventions and communicate design alternatives to the public \citep{simpson2001virtual, batty2001visualizing, billger2017search}. More recently, photomontage techniques and image compositing have been used to produce realistic-looking streetscape simulations from photographs, allowing planners to overlay proposed changes onto existing scenes \citep{sheppard2001guidance}. Despite these advances, such methods remain labor-intensive, require specialized expertise, and are difficult to scale across planning and design scenarios. Therefore, scholars have consistently called for more powerful tools to envision urban visuals for evidence-based scenario exploration or public participation \citep{batty2018inventing, steinitz2012framework}.


To visually represent urban environments, researchers have extensively leveraged street-view imagery (SVI) to evaluate streetscape quality by analyzing fine-grained visual elements observable at the human eye level \citep{dong2018green, gong2018mapping, zhang2019systematic}. Among the visual elements, the green view index \citep{yang2009can} has been widely adopted to measure visual exposure to vegetation and to reveal disparities in green space access across socioeconomic groups \citep{li2021examining, ki2021analyzing, sanchez_accessing_2024}. It has also been linked to physical activity and health outcomes, informing the design of healthier urban environments \citep{lu_using_2019, zhou_social_2019, ki2021analyzing}. In addition to greenery, sky-related metrics derived from SVIs - such as the sky view factor - have been used to evaluate urban density, ventilation potential, and heat resilience \citep{li_quantifying_2018}. This sky view metric has been operationalized to assess solar exposure and pedestrian thermal comfort in cities including Houston and Tokyo \citep{kim2023heat, chiang2023quantification, xia2021sky}. Built-form elements further contribute to the understanding of urban enclosure and perceived safety. Metrics such as the building view index, building edge density, and façade complexity have been employed to study aesthetic appeal, sense of security, and street-level economic activity \citep{naik2016cities, dubey2016deep, li2018investigating, wu_perceiving_2025}. Roads and sidewalks, as fundamental circulation elements, play a central role in analyses of accessibility, walkability, and traffic safety, where their visual presence, geometry, and layout are used to quantify connectivity and pedestrian friendliness \citep{tian2021assessing, li_measuring_2022, wei_innovative_2025}.

Beyond the visual elements, SVI has also been used to address broader social and sustainability questions. Prior work has inferred socioeconomic characteristics and inequalities from street-level imagery \citep{suel2019measuring}, evaluated landscape values and subjective perceptions through large-scale surveys linked to SVIs \citep{inoue_landscape_2022, ogawa_evaluating_2024, larkin_predicting_2021}, and identified disparities in neighborhood walkability using visually derived indices \citep{zhou_social_2019}. More recently, researchers have integrated SVI with complementary remote sensing data—such as satellite imagery—through advanced deep learning models to enhance predictive performance and robustness \citep{hu_identifying_2025}. Collectively, these studies demonstrate the critical role of SVI in advancing sustainable, equitable, and visually engaging built environments, extending its use beyond simple proxies of physical form \citep{kawakami_spatial_2013, biljecki_street_2021, han_gui_2024}. Nevertheless, the existing literature overwhelmingly focuses on measuring, evaluating, or predicting existing urban environments. Despite the rich semantic information embedded in SVIs, prior studies have rarely explored their potential for generative tasks, such as envisioning alternative or non-existing streetscapes conditioned on targeted design objectives, which remains a fundamental yet underexplored challenge.

To envision alternative or non-existing urban streetscapes, researchers must move beyond descriptive analysis and adopt generative AI techniques. Over the past decade, generative AI for image synthesis has evolved through several key stages. Early approaches, such as variational autoencoders (VAEs), introduced latent-variable models for image generation but often produced blurry and less realistic outputs \citep{ha2017neural}. Subsequent methods based on generative adversarial networks (GANs) significantly improved visual realism by learning through adversarial training \citep{larsen2016autoencoding, berthelot2018understanding}, yet their practical use was limited by training instability and sensitivity to data and model design \citep{saxena2021generative}. More recently, diffusion models have emerged as a new generation of generative AI, offering a more stable and flexible framework for high-quality image synthesis. Instead of generating images in a single step, diffusion models gradually transform random noise into realistic images through an iterative denoising process \citep{ho2020denoising, nichol2021improved}. This approach has been shown to produce visually coherent and semantically consistent results. Large-scale systems such as DALL·E, Imagen, and Stable Diffusion demonstrate the strong performance of diffusion models across diverse visual domains \citep{ramesh2022hierarchical, saharia2022photorealistic, rombach2022high}. 
At the same time, controllability has been substantially enhanced through models such as ControlNet, which allow users to guide image generation using visual constraints (e.g., semantic layouts or edge maps) without retraining the underlying model \citep{zhang_adding_2023}. Beyond general-purpose image synthesis, diffusion models have recently been applied to domain-specific visual generation tasks in architectural and urban contexts. In the architectural domain, diffusion-based methods have been explored for building façade generation and text-guided façade editing, demonstrating that ControlNet-based conditioning can produce structurally accurate and stylistically diverse architectural outputs \mbox{
\citep{ma2023text, brooks2023instructpix2pix}}\hskip0pt
. At the urban scene level, recent work has extended diffusion models to street-view synthesis conditioned on geometric layouts and bird's-eye-view maps, primarily motivated by data augmentation for autonomous driving perception \mbox{
\citep{gao2023magicdrive, deng_streetscapes_2024}}\hskip0pt
. Satellite-to-street-view synthesis has also been explored using cross-view diffusion frameworks \mbox{
\citep{li2024crossviewdiff,li2024sat2scene}}\hskip0pt
, demonstrating the potential of diffusion models to bridge different spatial scales of urban imagery. For geospatial analysis, the combination of textual and visual controls is especially powerful. Text prompts allow researchers to specify high-level design intentions, while visual constraints provide explicit spatial structure. Together, these capabilities position diffusion models as a promising tool for generating controllable and context-aware streetscape alternatives aligned with environmental objectives. 

Generative tasks are the regular practice of certain disciplines, such as urban planning and design, because practitioners in such disciplines need to visually envision alternative futures rather than merely evaluate existing conditions. Yet, despite decades of methodological advances, practitioners still lacked practical tools to translate abstract and text-based objectives directly into realistic streetscape visuals. Recent generative AI developments create a timely opportunity to close this gap. While early studies explored the generation of GIS maps or master plans using GANs or VAEs \citep{ye_masterplangan_2022, quan_artificial_2019, wu_ganmapper_2022}, these approaches remain limited in visual realism, stability, and controllability. Building on recent advances in diffusion models, this study introduces a controllable generative framework that synthesizes realistic SVI with explicitly specified visual goals. By aligning textual intentions with imagery constraints, our approach directly supports scenario exploration of streetscapes, bridging a long-standing divide between abstract textual objectives and visual street environments.

 \section{Data and Methods}
\label{sec:methods_data}

\subsection{Data collection and processing}
\label{sec:methods_data_collection}
\noindent This study focuses on two representative U.S. cities - Chicago and Orlando - which exhibit distinct urban morphologies and built environment patterns. 
Panoramic SVIs were systematically collected using \href{https://svd360.com/}{Street View Download 360} with regular 500-meter intervals to ensure spatially uniform coverage across diverse neighborhoods\footnote{This data collection approach has been widely used across past studies \citep{guo2025urban, tarkhan2025mapping}, although we also encourage future work to explore open-source SVIs such as Mapillary or KartaView.}. These images reflect the contrasting spatial forms and architectural typologies of the two cities: Chicago is characterized by dense, gridded streets and high-rise buildings, while Orlando features more dispersed development and low-rise suburban forms. The spatial boundary and representative images are presented in Supp Fig S1. 

We further transform the raw panoramic images into squared shapes. Each 360-degree panorama was first spatially mapped into four principal cardinal viewpoints—0°, 90°, 180°, and 270°—to simulate typical pedestrian perspectives. These projections were generated using a fixed horizontal field of view (FOV) of 120 degrees, which restricts the captured content to the forward-facing 120° cone in each direction. Additionally, the pitch angle was fixed at 90°, ensuring that the virtual camera looks straight ahead without tilting upward or downward. We split every projected image into left and right halves, yielding eight rectangular crops per panorama. Each crop was then resized to 640×640 pixels to match the model's input requirements. In total, the dataset includes 32,000 images covering a diverse built environment. This multi-perspective sampling strategy enriched the dataset by offering diverse ground-level visual representations of the urban environment. The segmentation approach also ensured that each location was captured from multiple angles, reducing the impact of visual obstructions such as buildings or vegetation, and enabling a more comprehensive analysis of urban features. The data processing details are visualized in Supp Fig 1.

We design textual descriptions for SVIs, which consist of three components: spatial context, segmentation values, and visual objects. The spatial context is described by the city name: \textit{The image is captured from Chicago/Orlando}. The segmentation values are captured as: \textit{The scene contains x$_1$\% road, x$_2$\% sidewalk, x$_3$\% building, x$_4$\% sky, x$_5$\% tree, and x$_6$\% person}. The percentages of segmentation labels are computed using the pre-trained SegFormer model with each image parsed into class-specific pixel ratios to represent the ratio of each segmentation label. Specifically, we use the ADE20K dataset which provides 150 semantic categories suited to diverse urban scenes, including tree, road, sky, and buildings, enabling finer-grained textual descriptions. The ADE20K dataset is substantially richer than alternative data sets, such as Cityscapes, which was designed primarily for autonomous driving perception and defines a relatively compact set of 30 classes for road-scene understanding. As a result, the visual objects are described as \textit{The image includes n$_1$ cars, n$_2$ persons, n$_3$ bicycles, and n$_4$ buses}. The object instances are identified using the pre-trained YOLOv11 model. 

\begin{figure}[htbp]
    \centering
    \includegraphics[width=1.0\linewidth]{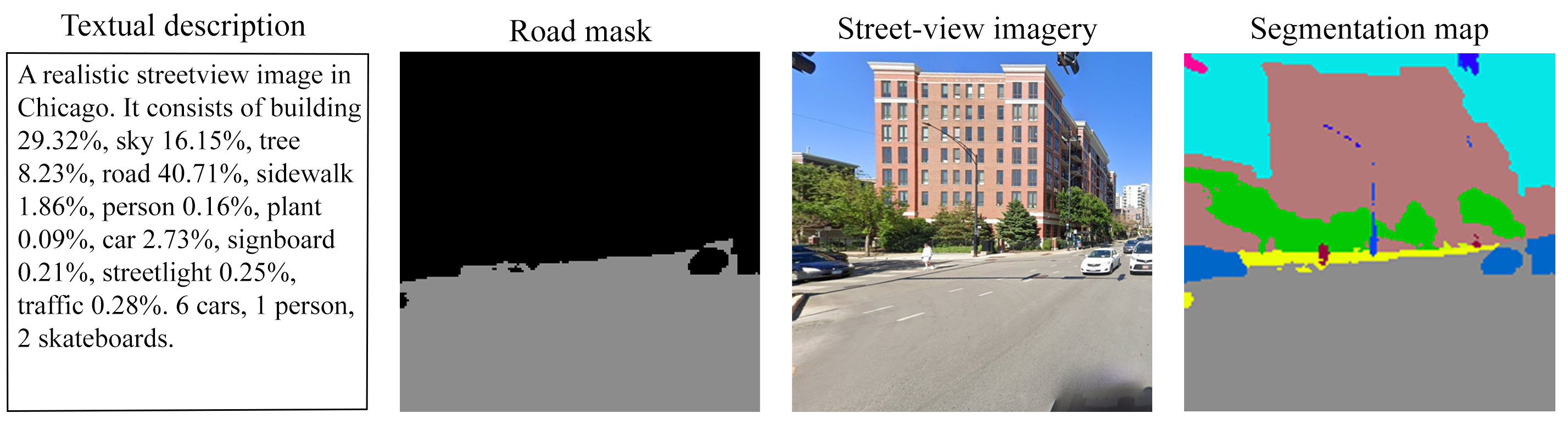}
    \caption{A data point from our multimodal dataset, including a textual description, a road mask, a street-view image, and a segmentation map. All the data modalities align because of our data processing pipeline.}
    \label{fig:data_example}
\end{figure}

Using semantic segmentation models, we derive both quantitative label proportions and pixel-level segmentation maps from street-view imagery. These segmentation maps serve dual roles as inputs for imagery control and as outputs for evaluation. In the simplest case, each SVI is segmented into road masks - typically fixed in urban development - and the surrounding built environment, which is more amenable to design modification. The resulting road masks are used as visual constraints to guide streetscape generation. For evaluation, generated SVIs are re-segmented to obtain label proportions, which are then compared with the target visual specifications encoded in the textual prompts to assess semantic consistency. An example data point is summarized in Figure~\ref{fig:data_example}. 

\subsection{Generative AI: Stable Diffusion and ControlNet}
\label{sec:methods_model}
\noindent The state-of-the-art diffusion models include Stable Diffusion (SD) and ControlNet. Stable Diffusion is a baseline model that enables imagery generation with textual controls, while ControlNet extends Stable Diffusion through both textual and imagery controls. Regarding the underlying mechanism, Stable Diffusion consists of a forward diffusion and a reversed diffusion process. In the forward diffusion process, Gaussian noise is progressively added to the image over $T$ steps, resulting in a noisy image denoted as $z_T$. In the reversed process, a neural network is trained to iteratively remove the noise $\epsilon$  and reconstruct the original image $z_0$ from the intermediate state $z_t$. New images are generated by leveraging the trained neural network in the reversed process. 

\begin{figure}[htbp]
    \centering
    \includegraphics[width=1\linewidth]{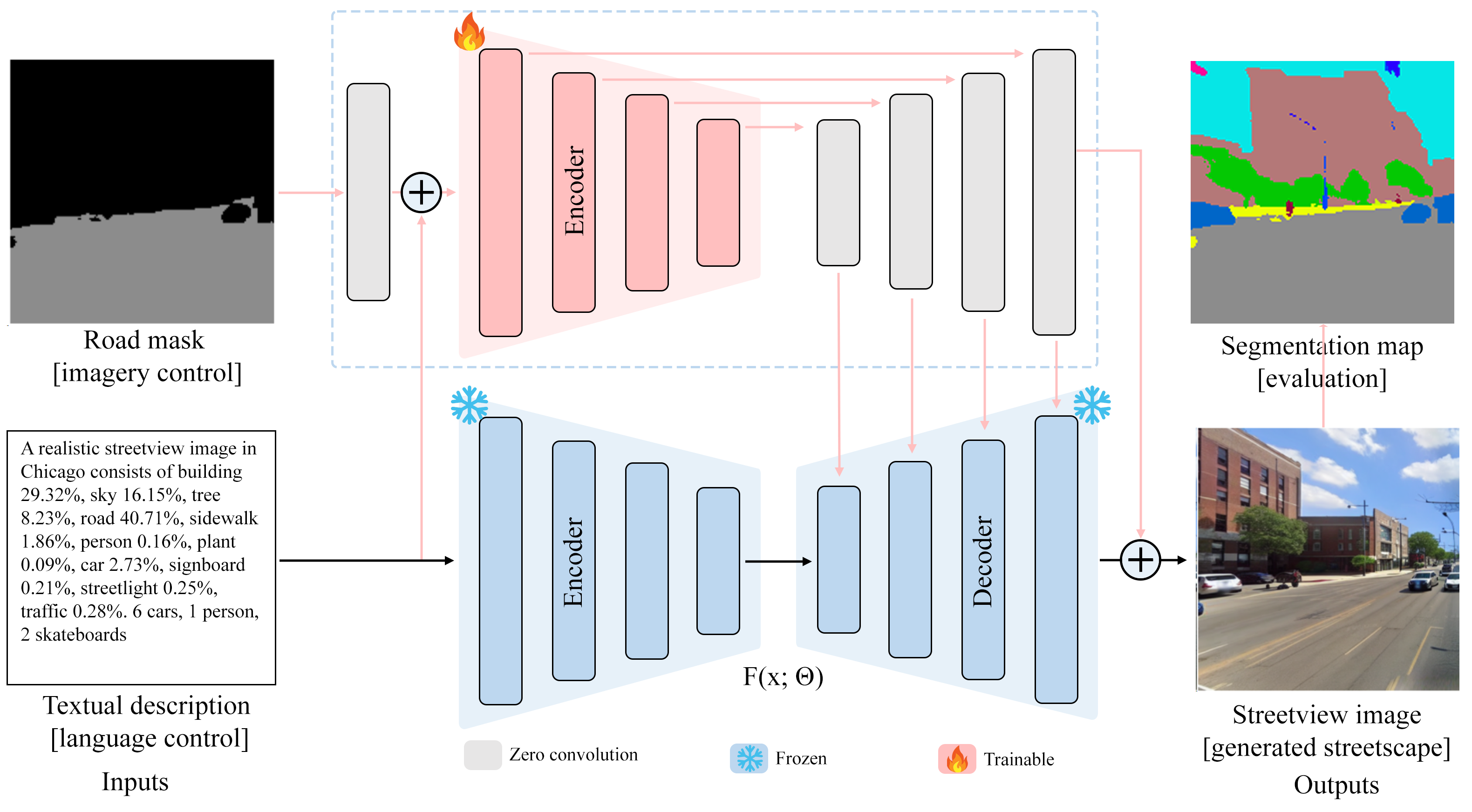}
    \caption{ControlNet architecture}
    \label{fig:controlnet}
\end{figure}

While being a powerful baseline, Stable Diffusion tends to generate too random images in layouts and label compositions because of its insufficient imagery controls. Therefore, we also introduce ControlNet, which strengthens Stable Diffusion through its imagery control as an extra input (Figure~\ref{fig:controlnet}). ControlNet includes the visual control map by leveraging a parallel set of trainable network blocks that operate alongside the frozen layers of the pretrained diffusion model. Mathematically, the frozen backbone of the pretrained diffusion model is denoted as $\mathcal{F}(x; \Theta)$, where $x$ is the textual description and $\Theta$ are fixed parameters. The parallel trainable branch is initialized using the same weights as the encoder: $\Theta_c = \Theta$. The visual control map, denoted as $c$, is first embedded via a zero-initialized $1 \times 1$ convolutional layer parameterized by $\Theta_{z1}$, and then combined with the textual description from the frozen branch. These modified inputs pass through the trainable branch, and finally yield the output image $y_c$, which equals to 
$
    y_c = \mathcal{F}(x; \Theta) + \mathcal{Z}\left(\mathcal{F}(x + \mathcal{Z}(c; \Theta_{z1}); \Theta_c); \Theta_{z2}\right).
$
To train the ControlNet, we need to predict the added noise at each diffusion step $t$ with the learning objective as:
\begin{equation}
    L(\Theta_c, \Theta_{z1}, \Theta_{z2}) = \mathbb{E}_{z_0, t, c, \epsilon \sim \mathcal{N}(0,1)} \left[ \left\| \epsilon - \epsilon_{\Theta_c, \Theta_{z1}, \Theta_{z2}}(z_t, t, c) \right\|^2 \right].
\end{equation}

As shown by Figure~\ref{fig:controlnet}, ControlNet can fully use our multimodal data points as its inputs and outputs. On the left hand side, the data with less information, such as road masks and textual descriptions, serves as the inputs. Such inputs are consistent with planning and design practices because urban planners typically design rich streetscape by starting with simple textual objects. On   the right hand side, the much richer data, i.e., SVI, serves as the outputs. This output also aligns with the fundamental goal in practice, as urban designers need to present the final visual solutions rather than some abstract numeric values. To evaluate the generated SVIs, we transform them into segmentation maps to compare to the initial textual descriptions, thus assessing the semantic consistency with textual descriptions and road masks.

 \subsection{Experiment design}
\label{sec:methods_experiment}
\noindent 
Our generative experiments investigate to what extent the generated streetscapes are realistic, controllable, and semantically consistent with the prespecified textual and visual controls. To achieve this goal, we conduct three groups of generative experiments. The first group focuses on establishing baseline generative results and testing the realism and semantic consistency with the textual controls extracted from the ground truth SVIs. The second group evaluates the effectiveness of textual controls beyond the observed values: we increase the greenery index or decrease sky view index significantly. For each variation, we observe whether the generated images reflect the expected compositional changes. The third group focuses on imagery controls, in which the textual prompt is held constant, but the road segmentation input varies. To compare the effectiveness of visual and language controls, we compare Stable Diffusion and ControlNet under identical training conditions. While both models receive the same textual prompts, only the latter incorporates the road segmentation map as the imagery control. This comparison enables us to evaluate whether visual control improves the visual realism and semantic consistency of generated streetscapes.

To evaluate the quality of streetscape generation, we adopt four common metrics from the generative AI literature, as summarized in Table~\ref{tab:evaluation_metrics}. These metrics evaluate \textit{global realism} - whether the generated imagery is overall realistic, and \textit{semantic consistency }- whether the generated imagery is consistent with the prespecified targets in textual and imagery controls. Specifically, the four evaluation metrics, include Fréchet Inception Distance (FID), Learned Perceptual Image Patch Similarity (LPIPS), Structural Similarity Index (SSIM), and mean Intersection over Union (mIoU). The FID metric compares the distributions of generated and real images, thus evaluating the realism of generated SVIs; LPIPS and SSIM directly assess perceptual and structural similarity between generated and ground truth SVIs; mIoU evaluates the pixel-wise consistency between the segmentation maps of the generated images and the original imagery masks, thus evaluating the consistency of the visual constraint with the generated SVI. Additionally, we include GPT-4o as an external commercial baseline, evaluated under identical textual prompts without road mask conditioning, to contextualize our results against general-purpose AI models.

\begin{table}[htbp]
\centering
\renewcommand{\arraystretch}{1.2}
\caption{Evaluation metrics assessing realism and consistency of generated SVIs.}
\label{tab:evaluation_metrics}

\begin{tabular}{p{3.2cm} p{7.2cm} p{3.0cm}}
\hline
\textbf{Metric} & \textbf{Comparison Objects} & \textbf{Range} \\
\hline
FID (Fréchet Inception Distance) &
Generated image vs. real image distribution to evaluate overall visual fidelity &
[0, $\infty$), lower is better \\
LPIPS (Learned Perceptual Image Patch Similarity) &
Generated SVI vs. ground truth SVI to assess perceptual similarity &
[0, 1], lower is better \\
SSIM (Structural Similarity Index) &
Generated SVI vs. ground truth SVI to measure structural consistency &
[0, 1], higher is better \\
mIoU (Mean Intersection over Union) &
Generated segmentation map vs. input segmentation map to evaluate semantic alignment &
[0, 1], higher is better \\
\hline
\end{tabular}
\end{table}

Our models are trained and tested on HiPerGator, the high-performance computing platform at the University of Florida, using NVIDIA A100 GPUs with 32 GB of memory. The training set consists of 29,000 SVIs, and the testing set includes 3,000 images held out for evaluation. The base Stable Diffusion model is fine-tuned using only textual descriptions, while the ControlNet is fine-tuned with both texts and road masks as inputs. Both models use a batch size of 8 and a learning rate of 0.00001. The Stable Diffusion model is trained for 22 epochs, consuming approximately 90 GPU hours, and the ControlNet model is trained for 17 epochs, taking about 72 GPU hours.

 \section{Results}
\label{sec:results}

\subsection{Generating benchmark streetscapes}
\label{sec:realistic_generation}
\begin{figure}[!tbp]
    \centering
    \includegraphics[width=1\linewidth]{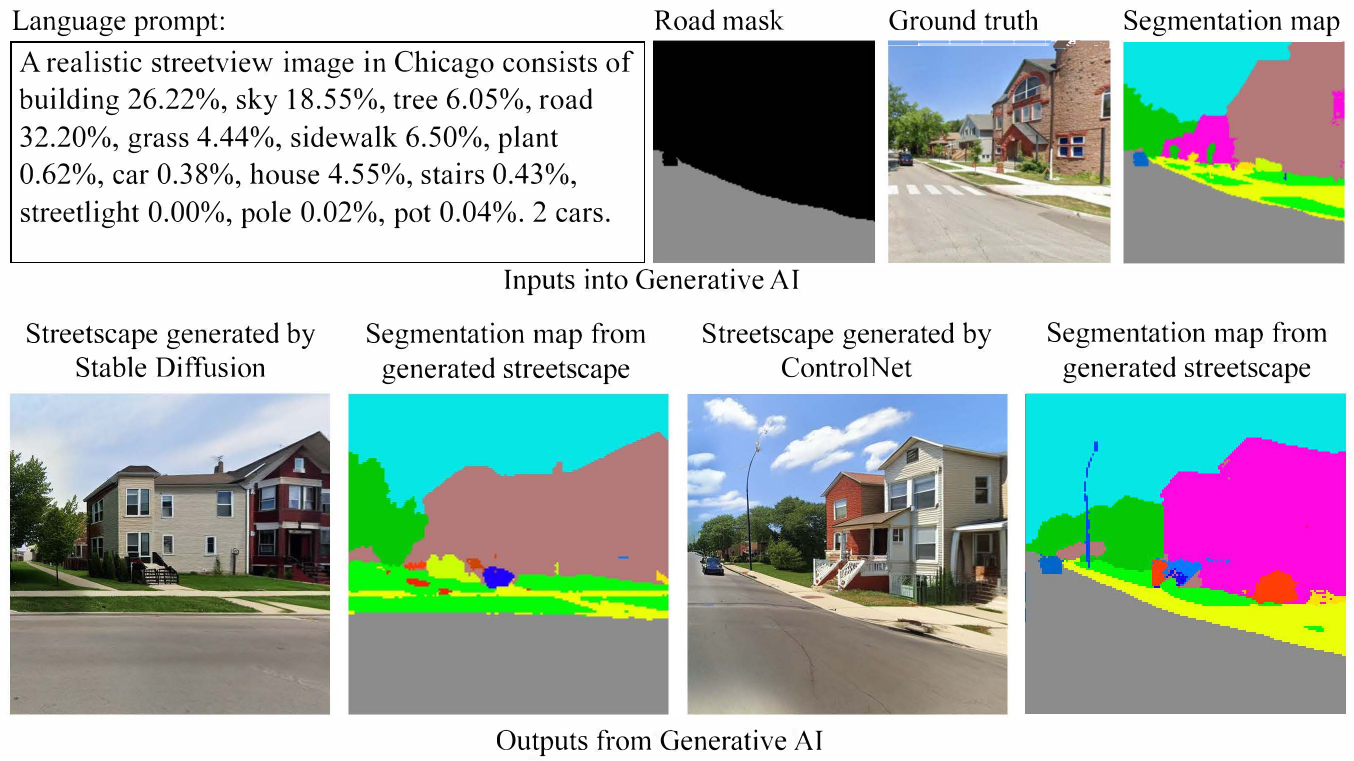} 
    \caption{Streetscape generation with Stable Diffusion and ControlNet
    }
    \label{fig:Generated images comparison}
\end{figure}

\noindent We first examine whether diffusion models can effectively generate realistic and semantically coherent street-view imagery using only textual descriptions extracted from the real SVIs. In the upper row, Figure~\ref{fig:Generated images comparison} visualizes the textual prompt, the road segmentation mask, a real SVI, and its corresponding segmentation map as the data inputs. In the lower row, Figure~\ref{fig:Generated images comparison} visualizes the generated streetscapes from Stable Diffusion and ControlNet, along with their corresponding segmentation maps. 

As shown in Figure~\ref{fig:Generated images comparison}, both Stable Diffusion and ControlNet can generate visually realistic streetscapes reflecting the input controls. Stable Diffusion can generate acceptable streetscapes because the outputs demonstrate reasonable spatial arrangements of roads, buildings, vegetation, and sky. For example, the key elements such as continuous roadways, vertical building edges, and urban greenery appear in spatially coherent positions. These results suggest that even without visual inputs, stable diffusion models can already capture the basic morphological logic of urban environments and render semantically meaningful streetscapes. ControlNet further improves visual realism and segmentation alignment because of the imagery control. As illustrated on the lower right side of Figure~\ref{fig:Generated images comparison}, ControlNet can successfully incorporate road segmentation masks and textual prompts as inputs, yielding streetscapes with clear road continuity, accurate building placement, and reduced visual artifacts. Compared to the baseline Stable Diffusion, ControlNet preserves the spatial structure of the road segmentation mask. 

Our quantitative evaluations further substantiate that ControlNet can visually improve   the semantic consistency of streetscape, as summarized in Table~\ref{tab:evaluation_results}. Compared to the baseline Stable Diffusion model without mask conditioning, ControlNet achieves consistently lower LPIPS scores across both cities, decreasing from 0.710 to 0.665 in Orlando and from 0.683 to 0.644 in Chicago, indicating an around 5$\sim$6\% improvement. Since LPIPS captures perceptual similarity, which measures the alignment between deep feature representations and human visual perception, such reductions indicate that streetscapes generated with road masks are perceptually closer to real street-view imagery, exhibiting more coherent scene composition. Improvements in semantic fidelity are further reflected in the mIoU scores, which increase from 0.156 to 0.193 (+23.72\%) in Orlando and from 0.153 to 0.224 (+46.41\%) in Chicago. The larger gain observed in Chicago suggests that explicit road structure guidance is particularly beneficial in dense urban environments, where complex spatial organization and strong interactions between road infrastructure and surrounding buildings pose greater challenges for purely textual control. These results demonstrate that ControlNet generation improves not only visual realism but also semantic alignment with real-world urban streetscapes. Notably, FID scores remain largely unchanged across Stable Diffusion and ControlNet, indicating that the incorporation of road segmentation masks does not alter the global image distribution at the dataset level. Instead, the improvements observed in LPIPS and mIoU suggest that ControlNet primarily enhances the visual quality of individual streetscapes while preserving overall realism. SSIM values exhibit only marginal variation, which is expected given their limited sensitivity to semantic correctness in complex urban scenes. 

\begin{table}[!htbp]
\centering
\renewcommand{\arraystretch}{1.2}
\caption{Quantitative evaluation results for generated images across two cities, comparing models with and without road segmentation mask guidance. Percentage changes relative to the corresponding w/o-mask baseline are reported in parentheses. Arrows indicate preferred metric direction.}
\label{tab:evaluation_results}

\begin{tabular}{lcccc}
\toprule
\textbf{Metrics} 
& \textbf{w/o mask Orlando} 
& \textbf{w/ mask Orlando} 
& \textbf{w/o mask Chicago} 
& \textbf{w/ mask Chicago} \\
\midrule
FID $\downarrow$   
& 63.98 
& 63.97 \,(-0.02\%) 
& 38.63 
& 38.62 \,(-0.03\%) \\

 LPIPS $\downarrow$ 
& 0.710 
& 0.665 \,(-6.34\%) 
& 0.683 
& 0.644 \,(-5.71\%) \\

 SSIM $\uparrow$  
& 0.199 
& 0.212 \,(+6.53\%) 
& 0.279 
& 0.273 \,(-2.15\%) \\

 mIoU $\uparrow$  
& 0.156 
& 0.193 \,(+23.72\%) 
& 0.153 
& 0.224 \,(+46.41\%) \\
mIoU(GPT-4o) $\uparrow$
& -- 
& 0.0058 
& -- 
& 0.0047 \\
 \midrule
\multicolumn{5}{l}{\textit{Class-wise mIoU for selected visual elements}} \\
\midrule
Tree $\uparrow$     
& 0.304 
& 0.408 \,(+34.21\%) 
& 0.269 
& 0.274 \,(+1.86\%) \\

 Sky $\uparrow$      
& 0.429 
& 0.426 \,(-0.70\%) 
& 0.504 
& 0.492 \,(-2.38\%) \\

 Building $\uparrow$ 
& 0.085 
& 0.174 \,(+104.71\%) 
& 0.089 
& 0.358 \,(+302.25\%) \\

 Road $\uparrow$     
& 0.350 
& 0.600 \,(+71.43\%) 
& 0.574 
& 0.742 \,(+29.27\%) \\
\bottomrule
\end{tabular}
\end{table}

As shown in Table~\ref{tab:evaluation_results}, class-wise mIoU results further demonstrate that visual controls can enhance the consistency between the targeted and the final view indices on generated SVIs. Road segments show the most pronounced gains, increasing from 0.350 to 0.600 in Orlando and from 0.574 to 0.742 in Chicago, confirming that the road mask effectively anchors the geometric backbone of the road view index from the streetscape. Importantly, improvements are also observed for buildings and vegetation, indicating that more accurate road geometry indirectly facilitates better placement and proportioning of urban view elements other than just road views. For reference, GPT-4o, a state-of-the-art general-purpose model, achieves mIoU scores of 0.0058 and 0.0047 in Orlando and Chicago respectively, substantially lower than our models. This confirms that domain-specific fine-tuning and spatial conditioning are essential for achieving semantic consistency in urban streetscape generation. Together, these results demonstrate that integrating road segmentation masks enables more faithful and realistic streetscape generation. 

\begin{figure}[htbp]
    \centering
    \includegraphics[width=1.0\linewidth]{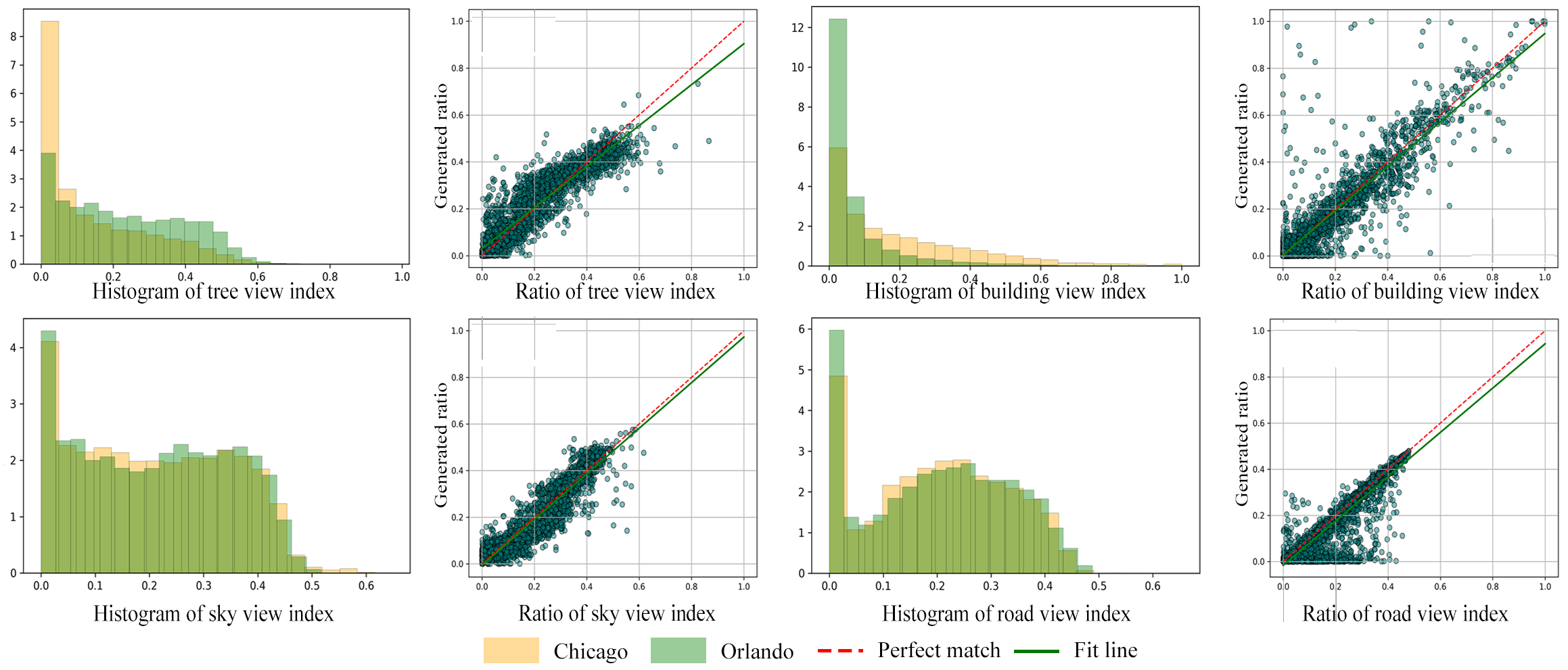}
    \caption{Evaluating the consistency between textual prompts and generated imagery. 
Histograms: Density distributions of tree, sky, building, and road proportions in generated streetscapes across Chicago and Orlando; 
Scatter Plots: Comparing the intended ratios of trees, sky, buildings, and roads from prompts with the actual proportions in generated images.}
    \label{fig:selected element}
\end{figure}

While Table~\ref{tab:evaluation_results} presents only the overall performance, Figure~\ref{fig:selected element} evaluates the semantic consistency between textual prompts and generated streetscape for individual data points. The histograms present density distributions of the proportions of trees, sky, buildings, and roads from the generated SVIs, while the scatter plots compare the intended ratios specified in the textual prompts with the actual proportions observed in the generated outputs. First, the density distributions reveal clear and interpretable urban patterns across cities. Chicago exhibits higher building view indices, consistent with its denser urban form, whereas Orlando shows relatively higher proportions of tree views. These distributions indicate that the generative model captures city-specific morphological characteristics rather than producing homogeneous or averaged streetscapes, demonstrating its ability to reflect distinct urban contexts. 

The scatter plots in Figure~\ref{fig:selected element} further demonstrate strong consistency between textual descriptions and generated streetscapes. Across all four elements - trees, sky, buildings, and roads - the majority of data points align closely with the diagonal, representing nearly perfect consistency between intended and generated proportions. All the fitted regression lines closely approximate this diagonal, indicating that ControlNet accurately translates textual specifications into corresponding visual outcomes. While buildings exhibit slightly greater dispersion than other elements, likely due to their complex geometries and variability in visual appearance, the overall alignment remains strong. Similarly, minor deviations for road proportions can be attributed to segmentation uncertainty in narrow street segments during evaluation. Importantly, these effects are limited in magnitude and do not undermine the overall agreement between prompts and generated scenes. Overall, this semantic consistency holds across both cities, highlighting the robustness of the approach under different urban conditions.

\subsection{Generating streetscapes by varying textual controls}
\label{sec:change_texts}
\noindent The section above uses the original textual descriptions extracted from the existing streetscapes. However, practitioners often need to create streetscapes based on new targets deviating from the existing environment. Therefore, to evaluate the controllability of our generative models, we conduct a series of experiments by adjusting the proportions of trees, sky, buildings, and roads, deviating further from the existing built environment. 

\begin{figure}[!htbp]
    \centering
    \includegraphics[width=0.75\linewidth]{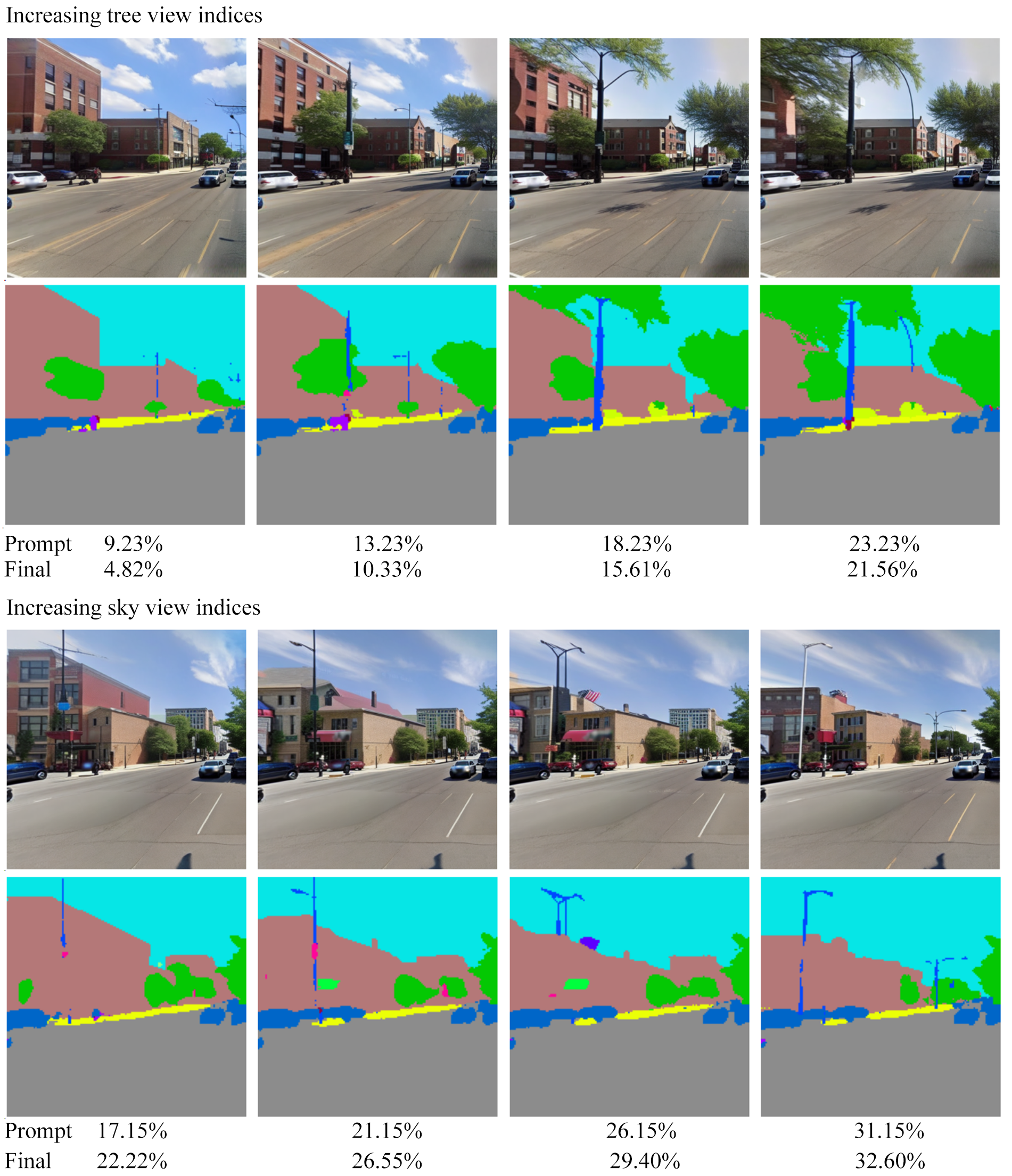}
    \caption{Increasing the proportion of green and sky view indices without changing the proportion of others}
    \label{fig:one_element_change}
\end{figure}

Figure~\ref{fig:one_element_change} demonstrates the controllability of the proposed generative framework by systematically varying individual visual elements in the textual prompts while holding others constant. Specifically, the figure illustrates how increasing the target proportions of tree and sky view indices leads to corresponding and spatially coherent changes in the generated streetscapes. In the top row, the tree view index is gradually increased in the textual prompts from 9.23\% to 23.23\%. The generated images and their corresponding segmentation maps show a consistent and monotonic increase in tree presence, with trees expanding along sidewalks and street edges while preserving reasonable locations relative to buildings and road infrastructure. Importantly, the generative experiment introduces additional greenery without encroaching on roads or distorting building geometry, indicating that changes in one visual element do not disrupt the overall scene structure. Similarly in the bottom row, increasing the sky view index from 17.15\% to 31.15\% produces a steady expansion of visible sky regions, primarily above rooftops and between buildings. The generated scenes exhibit smooth transitions in openness, maintaining realistic urban proportions and spatial relationships. The final generated sky proportions range from 22.22\% to 32.60\%, closely aligning with the target values specified in the prompts. Across both experiments, the close correspondence between intended and realized view indices highlights the sensitivity to fine-grained textual controls.

\begin{figure}[!htbp]
    \centering
  \includegraphics[width=0.75\linewidth]{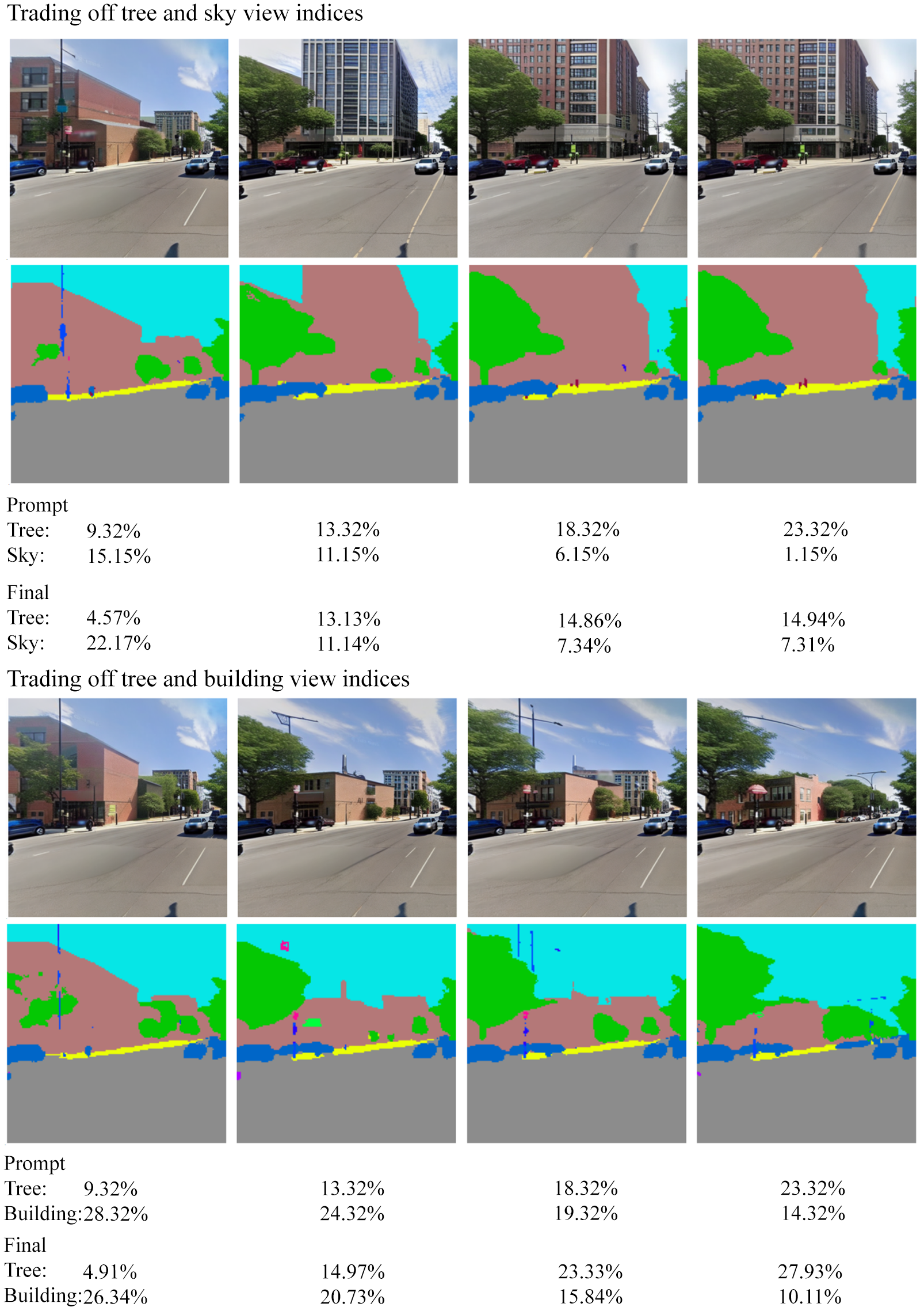}
     \caption{Increasing the proportion of tree view index while reducing the proportions of sky and buildings}
    \label{fig:two_element_change}
\end{figure}

Figure~\ref{fig:two_element_change} extends the analysis to multi-element control experiments, where two visual elements vary inversely within the same textual prompt. Specifically, the figure illustrates trade-offs between tree and sky view indices in the top row, and between tree and building view indices in the bottom row, representing more complex and realistic urban design interventions. In the top row, the tree view index is gradually increased from 9.32\% to 23.32\%, while the sky view index is reduced from 15.15\% to 1.15\% in the textual prompts. The generated streetscapes exhibit a clear and interpretable trade-off: as vegetation coverage expands along sidewalks and street edges, visible sky regions proportionally recede. The corresponding segmentation maps confirm this inverse relationship, with tree ratios increasing from 4.57\% to 14.94\% and sky ratios decreasing from 22.17\% to 7.31\%. These results indicate that the model accurately captures coupled constraints and responds coherently to compound prompt specifications. In the bottom row, a similar pattern is observed when trading off tree and building view indices. Increasing tree proportions from 9.32\% to 23.32\% results in a systematic reduction in building coverage, with generated building ratios decreasing from 26.34\% to 10.11\%. Importantly, the added vegetation replaces building mass in spatially plausible locations, such as street edges and foreground regions, without disrupting road geometry or overall scene layout. This behavior suggests that the model follows proportional cues embedded in the prompts while adhering to realistic urban form expectations. Across both scenarios, the generated streetscapes maintain contextual coherence despite the multi-dimensional constraints, demonstrating that ControlNet supports fine-grained, multi-dimensional control for streetscape generation. 

Beyond individual examples, we also assess the model’s responsiveness to textual control across all test samples, focusing on three dual-element trade-off scenarios: tree–sky, tree–building, and tree–road (Supp Fig S3). For each scenario, kernel density estimates summarize how the composition of generated images shifts in response to modified prompts that increase tree proportion while decreasing the paired element. In the tree–sky and tree–building scenarios, increasing the tree proportion in the language prompt leads to a clear rightward shift of the tree distribution by approximately 5\%, indicating that the generated images consistently reflect the intended compositional change. In the tree–road scenario, the increase in tree proportion remains evident, although the magnitude is slightly smaller than the targeted 5\%, reflecting the stabilizing influence of road infrastructure on scene composition. Despite these differences, all three scenarios respond clearly to textual prompts, demonstrating that ControlNet can accommodate prompt-driven trade-offs even when the specified visual proportions deviate from those in the original streetscapes. 

\subsection{Generating streetscapes by varying imagery controls and location contexts}
\label{sec:change_imagery}
\begin{figure}[thb!]
    \centering
    \includegraphics[width=.9\linewidth]{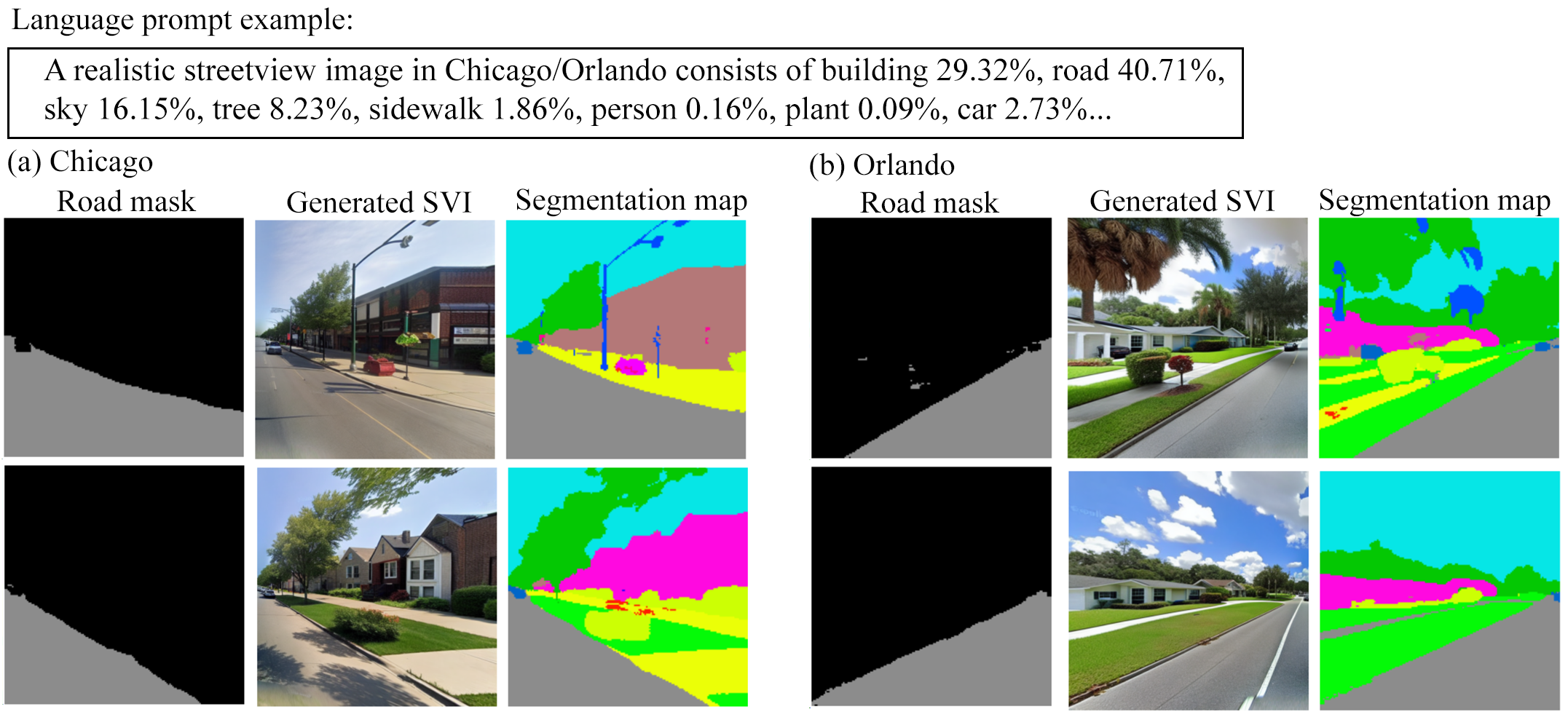}
    \caption{Generating street-view image with road mask and context-aware controls. (a) Context specified as Chicago with two road masks; (b) Context specified as Orlando with two road masks}
    \label{fig:context_road_control}
\end{figure}

\noindent To examine how diffusion models respond to other controlling approaches, we conduct an experiment where the language prompts are fixed while road segmentation maps and location contexts are systematically varied. 
Figure~\ref{fig:context_road_control} displays four experiments, including two from Chicago and two from Orlando. 
In all cases, the language prompt describes the target composition using consistent percentages for four major visual elements, and remains unchanged across rows. The only difference lies in the input road mask, which imposes a new visual constraint on the scene layout. 

Figure~\ref{fig:context_road_control} illustrates that both imagery and contextual controls are quite effective using the road segmentation masks and the city names in the language prompts. Across both Chicago and Orlando examples, the generated street-view images exhibit clear contextual consistency with their respective urban environments. In Chicago, the generated streetscapes reflect denser building frontages, narrower street corridors, and more enclosed urban canyons; whereas in Orlando, the streetscapes display wider roadways, open spatial layouts, and greater exposure to vegetation and sky. These differences emerge despite using the same generation framework, indicating that ControlNet effectively integrates location context alongside visual constraints. Meanwhile, the road segmentation masks serve as a strong and reliable visual control. In all cases, the generated streetscapes closely adhere to the geometry specified by the input road masks, including road width, curvature, and orientation. This semantic consistency between the generated imagery and the corresponding segmentation maps confirms that road infrastructure is consistently respected during generation, acting as a stable spatial backbone for the streetscape. Surrounding visual elements—such as buildings, trees, sidewalks, and sky—are adaptively arranged to fit the imposed road structure in a spatially coherent manner. Rather than distorting or overriding the road layout, these elements adjust naturally to the mask constraints, resulting in streetscapes that remain both visually plausible and semantically consistent. This semantic consistency is observed across different road configurations and across both cities, underscoring the robustness of road mask as imagery controls.

 \begin{figure}[!htbp]
    \centering
    \includegraphics[width=0.75\linewidth]{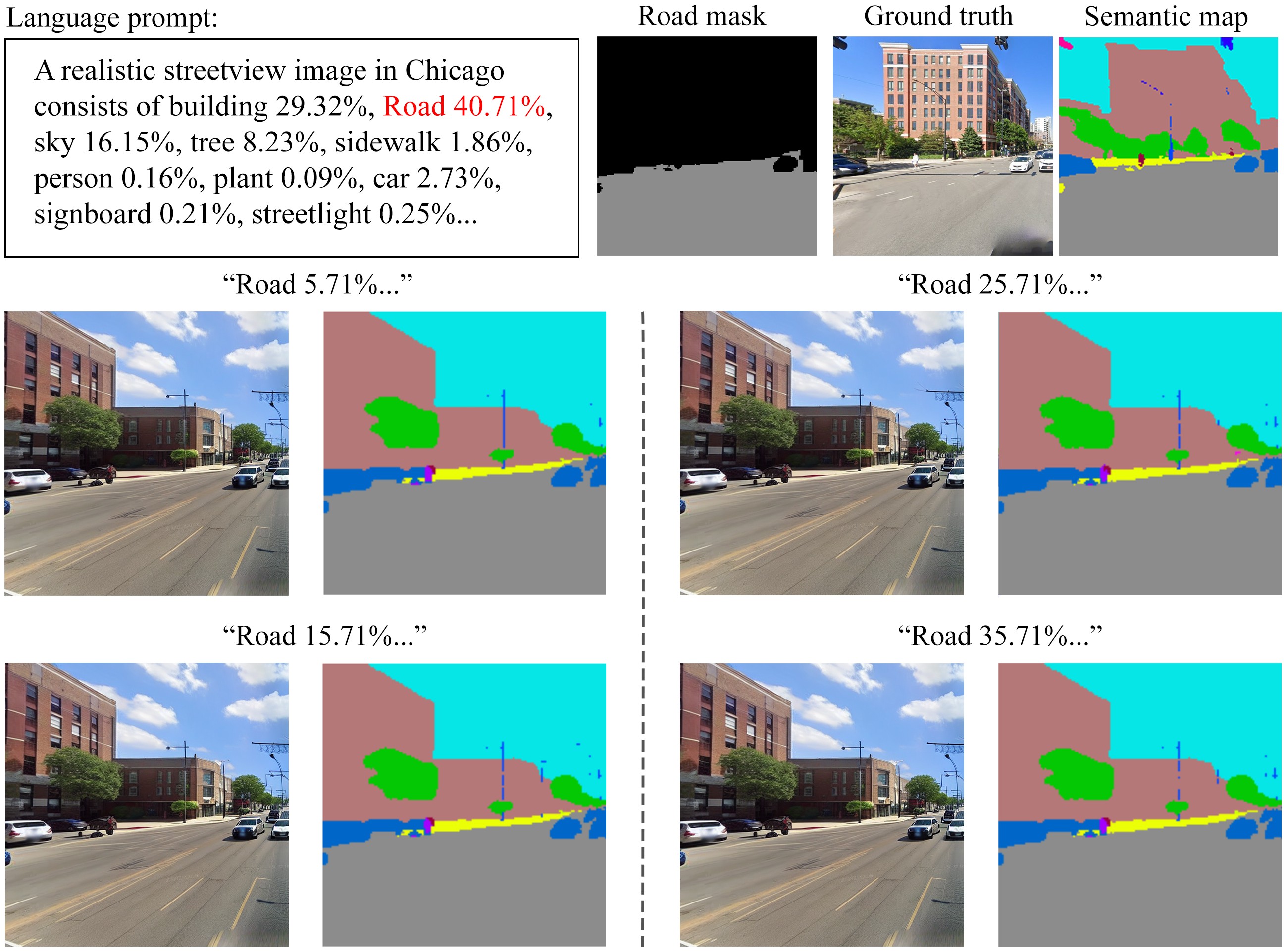}
    \caption{Streetscape generation by contrasting imagery and textual controls. The rows show generated images conditioned on increasing road proportions (5.71\% to 35.71\%) yet fixing the road mask, illustrating higher controllability using the imagery control.}
    \label{fig:comparing_text_visual_control}
\end{figure}

Figure~\ref{fig:comparing_text_visual_control} contrasts imagery and text controls under intentionally conflicting guidance, highlighting the relative strength of road segmentation masks as a more powerful control mechanism. In this controlled experiment, the road mask is fixed as the visual constraint, while the textual prompt specifies varying proportions of the road view index, ranging from 5.71\% to 35.71\%. Despite this six-fold variation in the textual specification of road proportion, the generated streetscapes and their corresponding segmentation maps remain highly consistent with the imposed road mask. Across all prompt settings, the segmentation results exhibit nearly identical road geometry, orientation, and coverage, closely matching the fixed road mask rather than reflecting the changing textual values. In other words, the realized road proportions in the generated images remain effectively unchanged, even when the prompt requests substantial decreases in road coverage. This outcome indicates that imagery controls exert a dominant influence on the process of streetscape generation whenever explicit spatial constraints are provided. Rather than attempting to reconcile conflicting instructions, the model prioritizes the segmentation-based road layout, preserving the structural backbone of the streetscape while limiting the impact of contradictory textual cues. Surrounding elements, such as buildings and vegetation, show only minor adjustments, further reinforcing the stability of the road masks as a visual guidance.

Importantly, this finding reflects a well-defined control hierarchy rather than a limitation of textual controls, and clarifies the complementary roles of the two input modalities. Spatial masks specify \textit{where} structural elements are located, providing explicit pixel-level constraints that anchor the scene's geometric backbone. Percentage-based textual prompts, by contrast, specify \textit{how much} of each visual element should appear globally, without prescribing spatial arrangement. It is worth noting that a semantic mask also implicitly encodes compositional information; for example, a road mask covering 40\% of the image already constrains scene proportions to a degree. In this case, numerical prompts may become partially redundant. However, textual prompts retain their practical value precisely in situations where no mask is available, such as during early-stage design exploration or when working in locations without pre-existing spatial data. In such cases, percentage-based prompts offer a lightweight and intuitive entry point for specifying compositional design intentions - increasing greenery, reducing building density - without requiring any pixel-level editing. Together, the two modalities address distinct practical needs: imagery controls for spatially precise generation, and textual prompts for high-level compositional guidance.

 \section{Conclusion}
\label{sec:conclusion}
\noindent Past studies have demonstrated the importance of SVIs in understanding urban environment. However, they rarely leveraged generative AI to envision alternative streetscapes conditioning on targeted objectives, which is an essential job for urban planners and designers. Here we introduce the state-of-the-art generative AI - diffusion models - to design alternative streetscapes by varying textual descriptions and visual inputs. By applying diffusion models to Chicago and Orlando, we have the following empirical findings.

Diffusion models can effectively synthesize realistic, semantically consistent, and controllable streetscape imagery by integrating textual and imagery controls. Through a series of controlled experiments across two contrasting urban contexts - high-density Chicago and lower-density Orlando - we systematically evaluated how different control modalities influence the global realism and semantic consistency of generated SVIs. Quantitative evaluations show that the imagery controls in ControlNet significantly improves semantic consistency. Compared to text-only controls, incorporating road segmentation masks reduces LPIPS by approximately 6\% in both cities, indicating stronger semantic alignment with real SVIs. More importantly, semantic consistency improves substantially, with overall mIoU increasing by 23.7\% in Orlando and 46.4\% in Chicago, and class-wise gains exceeding 70\% for road segments and 100–300\% for buildings. These results confirm that imagery controls enhance structural correctness without degrading global realism. Beyond aggregate metrics, we also demonstrate fine-grained controllability. By varying individual visual elements in textual prompts, ControlNet produces monotonic and spatially coherent changes in tree and sky view indices while preserving overall scene structure. More challenging multi-element trade-off experiments further show that the model can accommodate competing constraints, adjusting multiple elements in a coordinated manner. Lastly, the generative experiments contrasting textual and imagery controls reveal a clear control hierarchy. When textual prompts intentionally conflict with fixed road segmentation masks, the generated streetscapes consistently adhere to the visual constraints, even when the specified road proportion varies from 5.71\% to 35.71\%. Together, these empirical results establish a framework for the scenario planning of streetscape. By combining context-aware diffusion models with explicit structural controls, our approach enables realistic, semantically consistent, and controllable visualization of alternative urban scenarios, offering a promising tool for urban planning, design exploration, and human–AI co-creative workflows. 

Practitioners have long sought to create livable urban environment by directly translating abstract and text-based objectives into visual streetscapes. However, this goal was nearly impossible until the recent advancement of generative AI. Therefore, our research indicates transformative potentials for generative practices in the future. First, urban planners can use generative AI to visualize and compare alternative streetscapes before implementation. This approach aligns with long-standing calls for evidence-based and anticipatory planning~\citep{batty2018inventing, billger2017search}. For example, by specifying higher greenery ratios or modifying road segmentation inputs, planners can compare future streetscapes under various design goals. While this work is limited to only visual elements, future studies can incorporate more abstract goals such as reducing heat island effects, enhancing climate resilience, or improving perceived safety~\citep{li2022exploring, li2015does}. By translating abstract planning objectives to tangible streetscape visuals, generative models offer a novel interface between planning objectives and built environment. Second, generative models can enhance public engagement and participatory planning by translating complex planning objectives into street-view images. Traditional planning processes often rely on text-heavy documents or abstract diagrams, which are inaccessible to non-experts. In contrast, image-based representations grounded in real built environment constraints and policy prompts can make planning scenarios more understandable to residents and stakeholders~\citep{valencca2025creating}. This visual accessibility can enhance inclusive design discussions and enables meaningful community engagement during the planning process~\citep{guridi2025image}. Lastly, generative AI can potentially improve the productivity and creativity of urban planners. Rather than replacing professional expertise, generative models can complement human planners by producing diverse, context-aware streetscape alternatives. This function facilitates rapid ideation and accelerates the exploration of design possibilities, potentially leading to a new paradigm of co-creativity in planning~\citep{fu2024towards, burry2022new}. To make such scenario visualization accessible to non-technical users, future deployment could take the form of a lightweight web-based interface in which planners upload a road mask and adjust target visual proportions through intuitive sliders, with generated streetscapes returned in real time via cloud-hosted inference. Integration with existing GIS platforms such as ArcGIS or QGIS would further lower the adoption barrier by allowing road masks to be exported directly from familiar planning workflows. Moreover, it is possible to impose regulatory frameworks or design review workflows as the conditioning factors, beyond the simple textual and visual controls, which can further reduce repetitive drafting and enhance efficiency of urban planners. 


Despite these promising potentials, this study has several limitations that warrant further investigation. First, our experiments focus primarily on visual attributes observable in street-view imagery, such as greenery, sky, buildings, and road view indices, and have not yet incorporated non-visual or latent planning objectives, including socioeconomic factors, pedestrian comfort, traffic safety, or thermal performance. For example, the current framework does not explicitly encode formal urban design principles such as building setbacks or height-to-width ratios, as such annotations were not available in our training dataset. Future work could incorporate geometry-aware controls or regulatory constraints as additional conditioning factors to better align generated streetscapes with normative urban design standards. Second, while we evaluate two contrasting urban contexts, the generalizability of the findings to a broader range of cities remains an open question. Researchers can always expand the scope of two cities to more metropolitan areas in the US or worldwide. Third, the precision of percentage-based textual control is inherently constrained by the limited numerical discrimination of CLIP text embeddings, which tokenize percentage values as text strings and therefore lack an explicit sense of numerical magnitude or ordering. Future work could address this limitation by conditioning the diffusion model on raw scalar values. For example, feeding target proportions as continuous inputs through a dedicated lightweight encoder, rather than encoding them as text tokens, thus achieving finer-grained and more reliable quantitative control. Finally, practitioners should remain attentive to potential biases embedded in generative outputs. Models trained on existing imagery may inadvertently reflect local aesthetic norms from high-income communities, potentially misrepresenting low-income neighborhoods and distorting equity values in planning. Therefore, the generated outputs should be treated as exploratory scenario representations rather than prescriptive design solutions, and be supplemented with participatory design processes to ensure equitable and inclusive planning outcomes.


\printbibliography




\end{document}